\documentclass[conference]{IEEEtran}
\IEEEoverridecommandlockouts
\usepackage{cite}
\usepackage{amsmath,amssymb,amsfonts}
\usepackage{algorithmic}
\usepackage{graphicx}
\usepackage{textcomp}
\usepackage{xcolor}
\newtheorem{rem}{Remark}

\def\BibTeX{{\rm B\kern-.05em{\sc i\kern-.025em b}\kern-.08em
    T\kern-.1667em\lower.7ex\hbox{E}\kern-.125emX}}
\begin{document}

\title{Monte Carlo Functional Regularisation for Continual Learning\\
}

\author{\IEEEauthorblockN{Pengcheng Hao, Menghao Waiyan William Zhu, Ercan Engin Kuruoglu*}
\IEEEauthorblockA{\textit{Institute of Data and Information,}
\textit{Tsinghua Shenzhen International Graduate School,}
Shenzhen, China \\
pengchenghao@sz.tsinghua.edu.cn, zhumh22@mails.tsinghua.edu.cn, kuruoglu@sz.tsinghua.edu.cn}
}

\maketitle

\begin{abstract}
Continual learning (CL) is crucial for the adaptation of neural network models to new environments. Although outperforming weight-space regularisation approaches, the functional regularisation-based CL methods suffer from high computational costs and large linear approximation errors. In this work, we present a new functional regularisation CL framework, called MCFRCL, which approximates model prediction distributions by Monte Carlo (MC) sampling. Moreover, three continuous distributions are leveraged to capture the statistical characteristics of the MC samples via moment-based methods. Additionally, both the Wasserstein distance and the Kullback–Leibler (KL) distance are employed to construct the regularisation function. The proposed MCFRCL is evaluated against multiple benchmark methods on the MNIST and CIFAR datasets, with simulation results highlighting its effectiveness in both prediction accuracy and training efficiency.
\end{abstract}

\begin{IEEEkeywords}
Continual learning, Functional regularisation, Monte Carlo sampling,
\end{IEEEkeywords}

\section{Introduction}
\label{sec:intro}
Continual learning (CL)~\cite{CL-general-1,CL-general-2} refers to the ability of a neural network model to learn new tasks without forgetting previously acquired knowledge. This capability is especially valuable in dynamic environments where data evolves over time, eliminating the need to retrain models from scratch. For example, in healthcare, the CL supports the integration of new patient data into models without retraining, ensuring that the diagnostic systems stay current while safeguarding patient privacy~\cite{CL-app-1}. Also, the CL is essential for autonomous driving systems to adapt to dynamic environments, such as different weather conditions~\cite{CL-app-2}. Furthermore, a continual learning robot can continuously acquire new skills from unstructured real-world environments~\cite{CL-app-3}. However, CL methods suffer from catastrophic forgetting~\cite{CL-general-2}, where a model tends to lose previously learned knowledge when adapting to new tasks. 

To mitigate catastrophic forgetting, one presents weight-space regularisation CL methods, constraining the changes of model parameters during the acquisition of new knowledge. For instance, the Elastic Weight Consolidation (EWC)~\cite{EWC-1,EWC-2} leverages the Fisher information matrix to guide weight regularisation. Also, the synaptic intelligence (SI)~\cite{SI-1} constrains parameters according to their importance, which is evaluated via the entire training trajectory. Further, the CL method based on the Riemannian walk combines the regularization terms from SI and EWC, merging their respective advantages~\cite{RWALK}. By contrast, the variational continual learning (VCL)~\cite{VCL-1} approximates the weight posterior distributions by variational inference (VI)~\cite{VI-1,VI-2}. However, the complex relationship between model parameters and predictions makes the parameter-based regularisation ineffective in addressing catastrophic forgetting.

To address the limitations of weight regularisation, functional regularisation-based CL methods have been introduced, focusing on the intermediate or final output of neural networks. For instance, the functional regularisation for continual learning (FRCL)~\cite{FRCL} combines inducing point Gaussian process (GP)~\cite{GP} inference with deep neural networks but is limited to linear models. Also, the functional regularisation of the memorable past (FROMP)~\cite{FROMP} method utilises the Laplace approximation~\cite{Laplace-1} to estimate parameter variances, the optimisation of which is then not allowed. In contrast, the continual learning method via sequential function-Space variational inference (S-FSVI)~\cite{S-FSVI-1} approximates prediction distributions by linearisation of a Bayesian neural network (BNN)~\cite{BNN-1}, providing flexible optimisation over parameter means and variances. However, the S-FSVI requires high computational costs and large storage space due to the calculation of Jocabian matrices. Also, the adopted model linearisation introduces approximation errors.

In this work, we present a Monte Carlo (MC) sampling-based function-regularisation CL (MCFRCL) framework. The motivation for this choice is twofold. 1) The MC approach requires no Jacobian matrix computation and hence less computational loads. 2) BNNs are highly nonlinear systems, and compared with the linearisation method, the MC sampling can produce more precise uncertainty prediction. For instance, in nonlinear filtering, the classic extended Kalman filter~\cite{EKF} linearises the nonlinear system to estimate the prediction distribution. By contrast, the ensemble Kalman filter~\cite{EnKF} employs MC samples and achieves better estimation results. The contributions of this work consist of: 
\begin{enumerate}
    \item To approximate the intractable model prediction distributions, we first obtain prediction samples from the current model and the previous task model by MC sampling; 
    \item The prediction distributions are then approximated by three continuous densities, including Gaussian, Laplace and Cauchy~\cite{KL-Cauchy} distributions, the parameters of which are estimated by moment-based methods;
    \item To construct a regularisation function, the Wasserstein~\cite{wasserstein-1} and Kullback–Leibler (KL)~\cite{KL-1} distances are deployed to measure the similarity between the prediction distributions of the current and the previous models;
    \item In the simulation, the proposed method is compared with various weight/function-space regularisation-based methods on the MNIST and CIFAR datasets.
\end{enumerate}

The remainder of this paper is structured as follows: We begin, in Section~\ref{sec:Theoretical preliminary}, with an introduction to the theoretical background. Subsequently, Section~\ref{sec:MCFRCL} describes our proposed MCFRCL method, and the simulation results are elucidated in Section~\ref{sec:simulation}. Besides, Section~\ref{sec:conclusion} concludes this study.

\section{Theoretical preliminary}
\label{sec:Theoretical preliminary}
In this section, we introduce the employed 3 continuous distributions, followed by their corresponding moment-based parameter estimators and Wasserstein/KL distances. 
\begin{table}[h!]
\centering
\caption{Continuous distributions}
\begin{center}
\begin{tabular}{p{2cm} p{5.5cm}}
 \hline Notations & Definitions  \\
 \hline
 $\mathcal{N}(\mu,{\sigma}^2)$   & Univariate Gaussian pdf with mean  $\boldsymbol{\mu}$ and variance ${\sigma}^2$. \\ 
 $\mathcal{N}(\boldsymbol{\mu},\mathbf{\Sigma})$   & Multivariate Gaussian pdf with mean vector  $\boldsymbol{\mu}$ and covariance matrix $\mathbf{\Sigma}$. \\
 $\mathbf{Laplace}(a,b)$         & Univariate Laplace pdf with location $a$ and scale $b$. \\
  $\mathbf{Cauchy}(l,\gamma)$     & Univariate Cauchy pdf with location $l$ and scale $\gamma$.
 \\[1ex] 
 \hline
\end{tabular}
\end{center}
\label{table:distributions}
\end{table}

The definitions of the 3 densities are provided in Table~\ref{table:distributions}. Assume $x_{1:N} =\{x_\iota|\iota=1, \dots, N\}$ are $N$ samples from an univariate distribution. For the Gaussian distribution, its parameters can be estimated by the mean and variance of the samples. Similarly, the parameters of $\mathbf{Laplace}(a,b)$ and $\mathbf{Cauchy}(l,\gamma)$ can be estimated as follows:
\begin{equation}
\label{eq:moment estimation}
\begin{split}
\hat{a} &= \mathrm{mean}\left(x_{1:N}\right), \hat{b} = \sqrt{\mathrm{var}\left(x_{1:N}\right)/2}, \\
\hat{l} &= \mathrm{median}\left(x_{1:N}\right), \hat{\gamma} = \mathrm{mad}\left(x_{1:N}\right),      
\end{split}
\end{equation}
where the functions $\mathrm{mean}(.), \mathrm{var}(.), \mathrm{median}(.), \mathrm{mad}(.) $ represent the mean, variance, median, and median absolute deviation of the samples. Furthermore, the KL divergences between two univariate Gaussian~\cite{KL-Gaussian}, Laplace and Cauchy~\cite{KL-Cauchy} distributions, denoted as $\mathbb{D}_\mathrm{GKL}$, $\mathbb{D}_\mathrm{LKL}$, $\mathbb{D}_\mathrm{CKL}$, can be computed as follows:
\begin{equation}
\label{eq:distances-1}
\begin{split}
    \mathbb{D}_\mathrm{GKL}\left(p_1||p_2\right) &=  \mathrm{log}\frac{\sigma_2}{\sigma_1} + \frac{\sigma_1^2+\left(\mu_1-\mu_2\right)^2}{2\sigma_2^2}-\frac{1}{2} \\
    \mathbb{D}_\mathrm{LKL}\left(p_1||p_2\right) &=  \frac{b_1\mathrm{exp}\left(-\frac{|a_1-a_2|}{b_1}\right) +|a_1-a_2|}{b_2} +\mathrm{log}\frac{b_2}{b_1} -1\\
    \mathbb{D}_\mathrm{CKL}\left(p_1||p_2\right)& = \mathrm{log}\frac{(\gamma_1+\gamma_2)^2+(l_1-l_2)^2}{4\gamma_1\gamma_2}
\end{split}
\end{equation}
where $p_1,p_2$ are two univariate densities and their corresponding parameters have the same subscripts. Besides, in multivariate case, the square of Wasserstein distance~\cite{wasserstein-2} between two Gaussian distributions, $\mathbb{D}_\mathrm{GW}$, can be written as
\begin{equation}
\label{eq:distances-2}
\begin{split}
    \mathbb{D}_\mathrm{GW}^2\left(p_1,p_2\right) &=  ||\boldsymbol{\mu}_1-\boldsymbol{\mu}_2||_2^2 \\ &+\mathrm{Tr}\left(\mathbf{\Sigma}_1 +\mathbf{\Sigma}_2 - 2\left( \mathbf{\Sigma}_1^{1/2} \mathbf{\Sigma}_2 \mathbf{\Sigma}_1^{1/2}\right)^{1/2} \right). 
\end{split}
\end{equation}
In comparison, the Laplace and Cauchy distributions do not have a closed-form Wasserstein distance.

\section{MCFRCL}
\label{sec:MCFRCL}
\subsection{The proposed continual learning framework}
\label{sec:MCFRCL-1}
Assume a series of sequentially arriving tasks with index $t\in\{1,\dots,T\}$, where the dataset for the $t$-th task is denoted as $\mathcal{D}_t= \left(\mathbf{X}_t, \mathbf{y}_t \right)$. Considering a neural network $\mathbf{f}=f\left(.;\mathbf{\Theta}\right)$ with parameter $\mathbf{\Theta}$, the posterior estimation over the prediction function $\mathbf{f}$ can be expressed as 
\begin{equation*}
p\left(\mathbf{f}|\mathcal{D}_{1:t} \right) \propto p\left(\mathcal{D}_{t} | \mathbf{f} \right) p\left(\mathbf{f}|\mathcal{D}_{1:t-1} \right).
\end{equation*}
However, $p\left(\mathbf{f}|\mathcal{D}_{1:t} \right)$ is generally intractable. In~\cite{S-FSVI-1}, the sequential variational inference method is employed for the posterior approximation. Assume $q_t(\mathbf{f})$ is a functional variational distribution driven by a weight-space variational distribution $q_t(\mathbf{\Theta})$, the posterior approximation can be achieved by maximising
\begin{equation*}
 \mathcal{F}= \mathrm{E}_{q_t\left(\mathbf{f}\right)}\left[\mathrm{log}p\left(\mathcal{D}_{t}|\mathbf{f}\right)\right] - \mathbb{D}_{\mathrm{KL}}\left[ q_t\left(\mathbf{f}\right)||q_{t-1}\left(\mathbf{f}\right)\right], 
\end{equation*}
where $\mathcal{F}$ is the evidence lower bound (ELBO) and $\mathbb{D}_{\mathrm{KL}}\left[ q_t\left(\mathbf{f}\right)||q_{t-1}\left(\mathbf{f}\right)\right]$ is the function-space regularisation term.
Since there is no closed-form solution to $\mathcal{F}$, an approximation is proposed in~\cite{S-FSVI-1}: 
\begin{equation}
\label{eq:ELBO-1}
\begin{split}
 \mathcal{F} &\approx \frac{1}{S_\beta}\sum_{i=1}^{S_\beta} \mathrm{log} p\left( y_\beta | \mathbf{f}_i^{\beta}\right) \\
 &- \sum_{\tau=1}^{t-1}\sum_{k=1}^{D_\tau} \mathbb{D}_{\mathrm{KL}}\left[ q_t\left(\mathbf{f}^{\mathcal{C}_\tau}_k\right)||q_{t-1}\left(\mathbf{f}^{\mathcal{C}_\tau}_k\right)\right] 
 \end{split}
\end{equation}
where the prediction samples $\mathbf{f}_i^{\beta}=f\left(\mathbf{X}_\beta;\mathbf{\Theta}_i\right)$, $\mathbf{X}_\beta\in\mathbf{X}_t$ represents a batch of training data, $\mathbf{\Theta}_i$ is the $i$-th sample of $q_t(\mathbf{\Theta})$ and $S_\beta$ is the number of the prediction samples. Also, the number of model output dimensions for the $\tau$-th task is $D_\tau$. Besides, the output of the $k$-th dimension is $\mathbf{f}^{\mathcal{C}_\tau}_k= \left[f\left(\mathbf{X}^\mathcal{C}_\tau;\mathbf{\Theta}\right)\right]_k$, where the context set $\mathbf{X}^\mathcal{C}_\tau$ consists of $N_{\mathcal{C}_\tau}$ representative samples in the coreset of the $\tau$-th task and the coreset is sampled from the corresponding training dataset.  

Considering the benefits of both the KL and Wasserstein distance in functional regularisation~\cite{S-FSVI-1, wassertein-3}, we present our new objective function as
\begin{equation}
\label{eq:ELBO-2}
\begin{split}
 \mathcal{F} &\approx \frac{1}{S_\beta}\sum_{i=1}^{S_\beta} \mathrm{log} p\left( y_\beta | \mathbf{f}_i^{\beta}\right) \\
 &- \lambda \frac{N_\beta}{N_{\mathcal{C}_\tau}}\sum_{\tau=1}^{t-1}\sum_{k=1}^{D_\tau} \mathbb{D}\left[ q_t\left(\mathbf{f}^{\mathcal{C}_\tau}_k\right)||q_{t-1}\left(\mathbf{f}^{\mathcal{C}_\tau}_k\right)\right], 
\end{split}
\end{equation}
where $\mathbb{D} \in \{\mathbb{D}_\mathrm{GKL}, \mathbb{D}_\mathrm{LKL}, \mathbb{D}_\mathrm{CKL}, \mathbb{D}_\mathrm{GW}\}$. Also, $N_\beta$ is the batch size, $\frac{N_\beta}{N_{\mathcal{C}_\tau}}$ is used to alleviate the influence of the unbalanced data points for different tasks, and $\lambda$ is a scalar regularisation coefficient. However, the functional regularisation term in equation~(\ref{eq:ELBO-2}) is intractable, as there is no closed-form solution to the prediction distributions $q_t\left(\mathbf{f}^{\mathcal{C}_\tau}_k\right)$ and $q_{t-1}\left(\mathbf{f}^{\mathcal{C}_\tau}_k\right)$.

\subsection{Approximation of the regularisation function}
To handle the intractable prediction distribution, this section employs an MC sampling-based method to approximate the two variational distributions and then present an estimator for $ \mathbb{D}\left[ q_t\left(\mathbf{f}^{\mathcal{C}_\tau}_k\right)||q_{t-1}\left(\mathbf{f}^{\mathcal{C}_\tau}_k\right)\right]$ as shown in Figure~\ref{fig:Framework}. There are three steps:

\begin{figure}[h!]
\centering
\includegraphics[width=\linewidth]{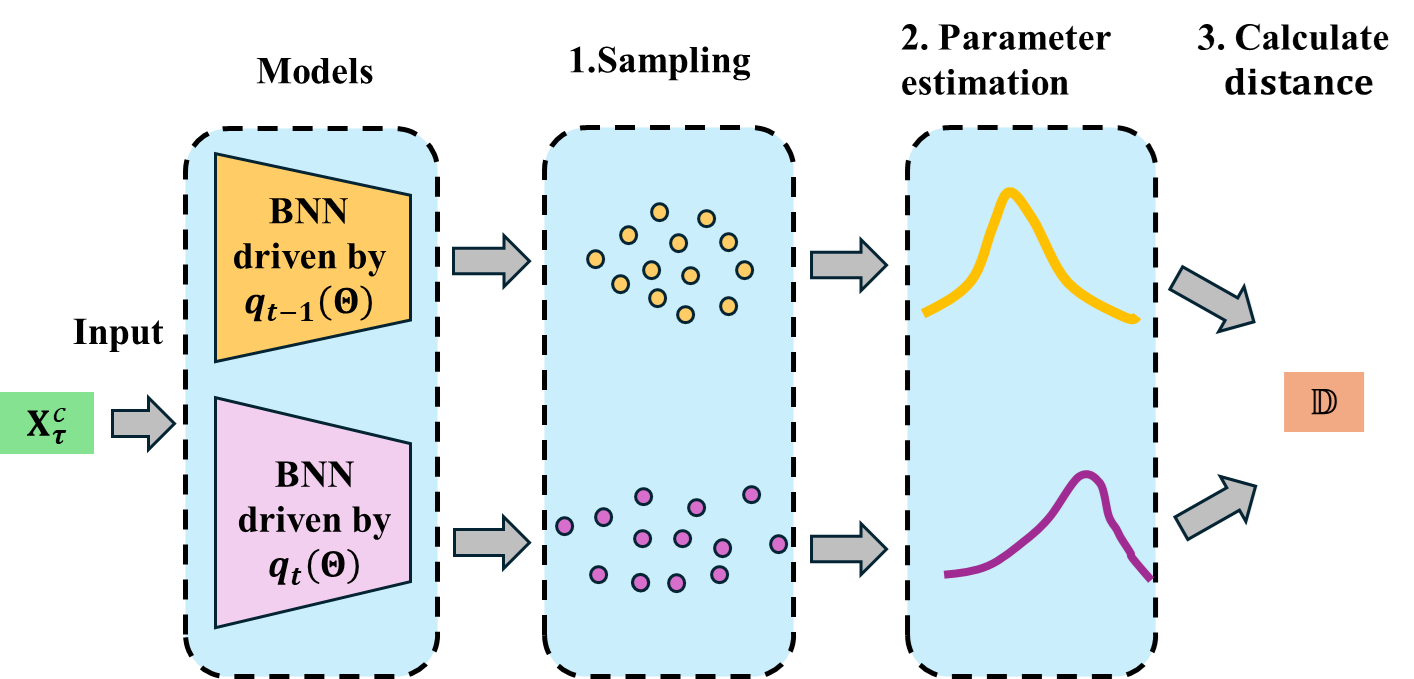}
\caption{The flowchart of the approximation method of the MCFRCL functional regularisation.}
\label{fig:Framework}
\end{figure}

\subsubsection{Monte Carlo sampling} With $\mathbf{X}^\mathcal{C}_\tau, \tau=1,\cdots, t-1$, we draw samples $\left[\mathbf{f}^{\mathcal{C}_\tau^+}_{k}\right]_j$ and $\left[\mathbf{f}^{\mathcal{C}^-_\tau}_{k}\right]_j$, $j=1,\dots,S_\mathcal{C}$, from $q_t\left(\mathbf{f}^{\mathcal{C}_\tau}_k\right)$ and  $q_{t-1}\left(\mathbf{f}^{\mathcal{C}_\tau}_k\right)$, respectively, where $S_\mathcal{C}$ is the number of samples. Also, $\left[\mathbf{f}^{\mathcal{C}_\tau^+}_{k}\right]_j=f\left(\mathbf{X};\mathbf{\Theta}_j^+\right)$, $\left[\mathbf{f}^{\mathcal{C}_\tau^-}_{k}\right]_j=f\left(\mathbf{X};\mathbf{\Theta}_j^-\right)$ where $\Theta_j^+ \sim q_t(\mathbf{\Theta})$ and $\Theta_j^- \sim q_{t-1}(\mathbf{\Theta})$. 

\subsubsection{Parameter estimation of the prediction distribution} We approximate $q_t\left(\mathbf{f}^{\mathcal{C}_\tau}_k\right)$ and  $q_{t-1}\left(\mathbf{f}^{\mathcal{C}_\tau}_k\right)$ by the continuous densities in Table~\ref{table:distributions}. To reduce the computational cost, we assume that all the components of $\mathbf{f}^{\mathcal{C}_\tau}_k$ are mutually independent, and hence the high-dimensional functional distribution approximation problem can be transformed into multiple one-dimensional estimation tasks. Given $\mathbf{f}^{\mathcal{C}_\tau}_{k,\xi}$, $\xi=1,\dots,N_{\mathcal{C}_\tau}$, is the $\xi$-th element of $\mathbf{f}^{\mathcal{C}_\tau}_{k}$, we have its corresponding one-dimensional samples  $\left[\mathbf{f}^{\mathcal{C}_\tau^+}_{k,\xi}\right]_j$ and $\left[\mathbf{f}^{\mathcal{C}^-_\tau}_{k,\xi}\right]_j$.  
Then we obtain the approximated univariate distributions $\hat{q}_t\left(\mathbf{f}^{\mathcal{C}_\tau}_{k,\xi}\right)$ and $\hat{q}_{t-1}\left(\mathbf{f}^{\mathcal{C}_\tau}_{k,\xi}\right)$, of which parameters can be estimated by~(\ref{eq:moment estimation}).

\begin{rem}
The adopted moment-based estimators in~(\ref{eq:moment estimation}) are computationally efficient and differentiable, which is beneficial to the model training process.     
\end{rem}

\subsubsection{Calculation of the distance $\mathbb{D}$} For $\mathbb{D} \in \{\mathbb{D}_\mathrm{GKL}$, $\mathbb{D}_\mathrm{LKL} $, $\mathbb{D}_\mathrm{CKL}\}$, as the KL distance is additive for independent distributions, we have 
\begin{equation*}
\begin{split}
 \mathbb{D}\left[ q_t\left(\mathbf{f}^{\mathcal{C}_\tau}_k\right) 
 ||q_{t-1}\left(\mathbf{f}^{\mathcal{C}_\tau}_k\right)\right] \approx 
\sum_{\xi=1}^{N_{\mathcal{C}_\tau}} \mathbb{D}  \left[ \hat{q}_t\left(\mathbf{f}^{\mathcal{C}_\tau}_{k,\xi}\right)||\hat{q}_{t-1}\left(\mathbf{f}^{\mathcal{C}_\tau}_{k,\xi}\right)\right].  
\end{split}
\end{equation*}
By contrast, for $\mathbb{D}_{\mathrm{GW}}$, we have
\begin{equation}
\label{eq:GW-1}
\begin{split}
 \mathbb{D}_{\mathrm{GW}}^2 & \left[ q_t\left(\mathbf{f}^{\mathcal{C}_\tau}_k\right)||q_{t-1}\left(\mathbf{f}^{\mathcal{C}_\tau}_k\right)\right] \approx \\
 &\qquad\qquad \sum_{\xi=1}^{N_{\mathcal{C}_\tau}} \mathbb{D}_{\mathrm{GW}}^2  \left[ \hat{q}_t\left(\mathbf{f}^{\mathcal{C}_\tau}_{k,\xi}\right)  ||\hat{q}_{t-1}\left(\mathbf{f}^{\mathcal{C}_\tau}_{k,\xi}\right)\right],  
\end{split}
\end{equation}
which can be easily derived from~(\ref{eq:distances-2}).


\subsection{Discussion}
Our proposed method is mostly relevant to the function-space regularisation methods, including FRCL~\cite{FRCL}, FROMP~\cite{FROMP}, and S-FSVI~\cite{S-FSVI-1}. The FRCL only treats the weights of the last layer in a neural network as random, whilst our method is applicable to fully stochastic models. Also, compared with the FROMP, the MCFRCL achieves an optimisation on both means and variances of parameters. Furthermore, unlike the S-FSVI, which relies on linear approximations, the MCFRCL utilises MC sampling. This approach avoids the expensive computations of Jocabian matrices and generally provides more accurate predictions, especially in highly nonlinear systems.

\section{Empirical Evaluation}
\label{sec:simulation}
In this section, we evaluate our proposed MCFRCL method.
Section~\ref{sec:MNIST data} and~\ref{sec:CIFAR data} introduce the CL tasks based on the (Fashion) MNIST and CIFAR datasets, respectively. In Section~\ref{sec:performance MNIST} and~\ref{sec: CIFAR exp}, various CL approaches are compared with four MCFRCL variants based on $\{\mathbb{D}_\mathrm{GKL}, \mathbb{D}_\mathrm{LKL}, \mathbb{D}_\mathrm{CKL}, \mathbb{D}_\mathrm{GW}\}$, with benchmark results directly sourced from~\cite{S-FSVI-1} and~\cite{S-FSVI-2} to ensure strong baselines.

\subsection{Split (Fashion) MNIST setup}
\label{sec:MNIST data}
1) Split (Fashion) MNIST comprises five tasks, each involving binary classification between a pair of (Fashion) MNIST classes. Also, both the MNIST and Fashion MNIST datasets contain 60,000 samples for training and 10,000 samples for testing, and all images are transformed into floating-point numbers ranging from 0 to 1. 2) We employ single-head fully connected neural networks with two hidden layers of size 256, applying the ReLU activation function to all non-output units. Besides, an Adam optimiser with an initial learning rate of $0.0005$ $(\beta_1 = 0.9; \beta_2 = 0.999)$ is adopted. 3) For the split MNIST (S-MNIST) experiment, we evaluate two scenarios with 40 and 200 coreset points per task, using 10 and 60 epochs, respectively. By contrast, for the split Fashion MNIST (S-FMNIST) tasks, the coreset size is manually set to 200 points per task, with 5 epochs per task. Additionally, during training on the first task, context points are generated by sampling each pixel uniformly from the range $[0,1]$. For subsequent tasks, context points are randomly selected from the coreset. 4) For the first task, we assume a Gaussian functional prior distribution with zero mean and a diagonal covariance of magnitude $0.001$.  5) Set $N_\beta$=128, $N_{\mathcal{C}_\tau}$=40, $S_\beta$=30,  $S_\mathcal{C}$=30. Also, $\lambda$ is selected from $\{10^n, 3 \times 10^n \mid n \in[-9,7], n \in \mathbb{Z}\}$, and we set the optimal result as the final result. Besides, all (Fashion) MNIST experiments are conducted with 10 MC runs.

\subsection{Split CIFAR  setup}
\label{sec:CIFAR data}
1) Split CIFAR~\cite{FROMP} comprises six tasks. The first involves ten-way classification using the entire CIFAR-10 dataset, while each of the remaining five tasks also involves ten-way classification with classes selected from CIFAR-100. 2) As in~\cite{S-FSVI-1}, we utilize a neural network consisting of four convolutional layers, followed by two fully connected layers and multiple output heads, one for each task. Also, we use the same Adam optimiser as in the Split (Fashion) MNIST setup. 3) The coreset size is fixed at 200, with 120 epochs for the first task and 50 epochs for the subsequent tasks. Also, for the first task, context points are generated by sampling each pixel uniformly from the range $[0,1]$, while context points are randomly selected from the coreset in the subsequent tasks. 4) For the first task, the functional prior distribution is assumed to be Gaussian with zero mean and a diagonal covariance of magnitude 1.0. 5) Set $N_\beta$=512, $N_{\mathcal{C}_\tau}$=50, $S_\beta$=5,  $S_\mathcal{C}$=30. Also, $\lambda$ is selected from $\{10^n \mid n \in[-9,2], n \in \mathbb{Z}\}$, and we set the optimal result as the final result.  Besides, 5 MC runs are given for all CIFAR experiments.

\subsection{Performance on the MNIST dataset}
\label{sec:performance MNIST}
In this experiment, the MCFRCL variants are compared with various benchmark methods, including Online EWC~\cite{EWC-2}, SI~\cite{SI-1}, VCL~\cite{VCL-1}, VAR-GP~\cite{VAR-GP}, FROMP~\cite{FROMP}, S-FSVI~\cite{S-FSVI-1}. The prediction accuracy comparison is shown in Table~\ref{table:MNIST benmarks}, and the bold numbers highlight the best results. The optimal MCFRCL consistently outperforms the other benchmark methods across all scenarios. Also, the optimal MCFRCL variant varies depending on the datasets, while the MCFRCL with $\mathbb{D}_\mathrm{CKL}$ produces worse estimation than the other variants. This indicates that the heavy-tailed Cauchy distribution fails to accurately capture the prediction uncertainty.

\begin{table}[h]
\centering
\caption{Comparison of prediction accuracy ($\%$) on MNISTs}
\begin{center}
\begin{tabular}{c c c c} 
 \hline Method & S-MNIST & S-MNIST  & S-FMNIST  \\
      & 40 pts/task  & 200 pts/task  & 200 pts/task
 \\ 
 \hline
 Online EWC  & 19.95±0.28 & 19.95±0.28 & 19.95±0.28 \\ 
 SI & 19.82±0.09 & 19.82±0.09 & 19.80±0.21 \\
 VCL & 22.31±2.00 & 32.11±1.16 & 53.59±3.74 \\
 VAR-GP  & - & 90.57±1.06 & - \\
 FROMP & 75.21±2.05 & 89.54±0.72 & 78.83±0.46 \\ 
 S-FSVI   & 84.51±1.30 & 92.87±0.14 & 77.54±0.40 \\
 \hline
 MCFRCL:  &  &  &  \\
 $\mathbb{D}_\mathrm{GW}$  & 83.30±1.82 & \textbf{93.22±0.39} & 78.3±4.33 \\
 $\mathbb{D}_\mathrm{GKL}$  & 82.43±1.31 & 92.63±0.44 & \textbf{79.15±1.02} \\
 $\mathbb{D}_\mathrm{LKL}$  & \textbf{84.85± 0.88} & 92.88±0.42 & 73.76±6.23 \\
 $\mathbb{D}_\mathrm{CKL}$  & 65.84±5.77 & 89.59±1.51 & 33.14±5.67 \\[1ex] 
 \hline
\end{tabular}
\end{center}
\label{table:MNIST benmarks}
\end{table}

\begin{figure}[h!]
\centering
\includegraphics[width=\linewidth]{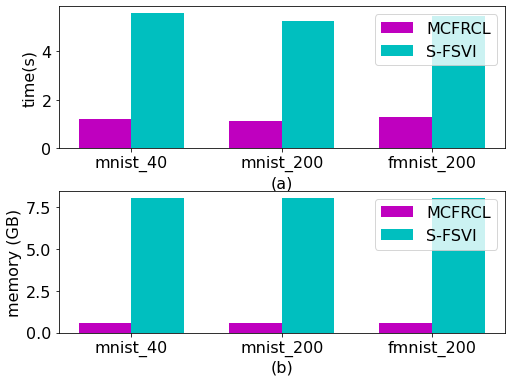}
\caption{Average training time per epoch and occupied GPU memory during the last task.}
\label{fig:time memory}
\end{figure}

Also, Figure~\ref{fig:time memory} presents the average training time per epoch and required GPU memory of the optimal MCFRCL variant and the S-FSVI. In all 3 scenarios, the MCFRCL requires less training time and GPU memory as the S-FSVI suffers from the computationally expensive Jacobian matrix.

Besides, the influence of the MC sample sizes $S_\mathcal{C}$ and $S_\beta$ are illustrated in Table~\ref{table:Ablation MC samples}. Due to the computational cost, our method is not scalable to large sample sizes. From Table~\ref{table:Ablation MC samples}, the small changes in sample sizes have little influence on model performance.

\begin{table}[h]
\centering
\caption{Comparison of Prediction accuracy ($\%$) with different numbers of MC samples on FMNIST}
\begin{center}
\begin{tabular}{c c c c c} 
 \hline  &  & $S_\mathcal{C}$=5 & $S_\mathcal{C}$=30  & $S_\mathcal{C}$=100  \\
 \hline
 $\mathbb{D}_\mathrm{GW}$ & $S_\beta$=10  & 79.02±3.81 & 77.50±4.68 & 78.08±4.12 \\
  & $S_\beta$=30  & 79.96±3.00 & \textbf{80.12±2.95} & 78.54±4.55 \\
 & $S_\beta$=100  & 78.21±4.44 & 77.33±4.60 & 79.01±3.83 \\
 \hline
 $\mathbb{D}_\mathrm{GKL}$ & $S_\beta$=10  & 79.25±2.67 & 78.33±3.05 & 78.05±2.85 \\
  & $S_\beta$=30  & 79.22±2.77 & \textbf{79.33±0.70} & 78.35±2.78 \\
 & $S_\beta$=100  & 78.95±2.90 & 78.30±2.90 & 78.01±3.05 \\
 \hline
\end{tabular}
\end{center}
\label{table:Ablation MC samples}
\end{table}

\subsection{Performance on the CIFAR dataset}
\label{sec: CIFAR exp}
The performance of the MCFRCL on the CIFAR dataset is presented in Table~\ref{table:CIFAR benmarks}. There are 2 scores, the prediction accuracy and backward transfer (BT), and higher values are better for both. The bold numbers indicate the best results. Our proposed method produces more accurate predictions than the weight-regularisation CL methods, including the EWC and the VCL. However, compared with the functional regularisation-based benchmarks, the MCFRCL performs worse, which suggests the limitations of the MC sampling in capturing complex statistical characteristics of large stochastic models. Also, the MCFRCL has lower BT scores than the other methods, which implies that it suffers serious catastrophic forgetting in this scenario.
\begin{table}[h!]
\centering
\caption{Performance comparison on CIFAR Datasets}
\begin{center}
\begin{tabular}{c c c c} 
 \hline Method & Accuracy($\%$)  & BT
 \\ 
 \hline
 EWC  & 71.6±0.4 & \textbf{-2.3±0.6} \\ 
 VCL & 67.4±0.6   & -9.2±0.8 \\
 FROMP & 76.2±0.2  & -2.6±0.4 \\ 
 S-FSVI & \textbf{77.6±0.2} & -2.5±0.2 \\
 \hline
 MCFRCL:  &  &  &  \\
 $\mathbb{D}_\mathrm{GW}$    & 73.08±0.2 & -7.99±0.4 \\
 $\mathbb{D}_\mathrm{GKL}$   & 72.49±0.2   & -8.29±0.2 \\
 $\mathbb{D}_\mathrm{LKL}$   & 73.16±1.1   & -8.43±0.7 \\
 $\mathbb{D}_\mathrm{CKL}$   & 67.8±0.5  &  -14.47±0.5 \\[1ex] 
 \hline
\end{tabular}
\end{center}
\label{table:CIFAR benmarks}
\end{table}
\section{Conclusions}
\label{sec:conclusion}
In this work, we introduce the MCFRCL, a novel functional regularisation-based continual learning framework, where three continuous distributions approximate the model prediction distributions via MC sampling and moment-based methods. Also, both the Wasserstein and KL distances are deployed to construct the regularisation function. Our proposed method is compared with various benchmark CL frameworks. The experiment results demonstrate that the MCFRCL achieves better prediction accuracy and training efficiency on MNIST datasets. By contrast, on the more complicated CIFAR dataset, while outperforming the weigh-regularisation methods, the MCFRCL falls short compared to other function-regularisation benchmarks. This suggests that the employed MC sampling is ineffective at approximating complex model prediction densities with a small number of samples. In the future, we will consider applying our proposed method to some edge devices, such as health monitoring systems~\cite{edge_device}, which require light and fast models due to the limited computational power.

\bibliographystyle{IEEEtran}
\bibliography{mybibfile}

\begin{thebibliography}{10}
\providecommand{\url}[1]{#1}
\csname url@samestyle\endcsname
\providecommand{\newblock}{\relax}
\providecommand{\bibinfo}[2]{#2}
\providecommand{\BIBentrySTDinterwordspacing}{\spaceskip=0pt\relax}
\providecommand{\BIBentryALTinterwordstretchfactor}{4}
\providecommand{\BIBentryALTinterwordspacing}{\spaceskip=\fontdimen2\font plus
\BIBentryALTinterwordstretchfactor\fontdimen3\font minus \fontdimen4\font\relax}
\providecommand{\BIBforeignlanguage}[2]{{%
\expandafter\ifx\csname l@#1\endcsname\relax
\typeout{** WARNING: IEEEtran.bst: No hyphenation pattern has been}%
\typeout{** loaded for the language `#1'. Using the pattern for}%
\typeout{** the default language instead.}%
\else
\language=\csname l@#1\endcsname
\fi
#2}}
\providecommand{\BIBdecl}{\relax}
\BIBdecl

\bibitem{CL-general-1}
L.~Wang, X.~Zhang, H.~Su, and J.~Zhu, ``A comprehensive survey of continual learning: theory, method and application,'' \emph{IEEE Transactions on Pattern Analysis and Machine Intelligence}, 2024.

\bibitem{CL-general-2}
M.~De~Lange, R.~Aljundi, M.~Masana, S.~Parisot, X.~Jia, A.~Leonardis, G.~Slabaugh, and T.~Tuytelaars, ``A continual learning survey: Defying forgetting in classification tasks,'' \emph{IEEE transactions on pattern analysis and machine intelligence}, vol.~44, no.~7, pp. 3366--3385, 2021.

\bibitem{CL-app-1}
T.~Verma, L.~Jin, J.~Zhou, J.~Huang, M.~Tan, B.~C.~M. Choong, T.~F. Tan, F.~Gao, X.~Xu, D.~S. Ting \emph{et~al.}, ``Privacy-preserving continual learning methods for medical image classification: a comparative analysis,'' \emph{Frontiers in Medicine}, vol.~10, p. 1227515, 2023.

\bibitem{CL-app-2}
E.~Verwimp, K.~Yang, S.~Parisot, L.~Hong, S.~McDonagh, E.~P{\'e}rez-Pellitero, M.~De~Lange, and T.~Tuytelaars, ``Clad: A realistic continual learning benchmark for autonomous driving,'' \emph{Neural Networks}, vol. 161, pp. 659--669, 2023.

\bibitem{CL-app-3}
S.~Auddy, J.~Hollenstein, M.~Saveriano, A.~Rodr{\'\i}guez-S{\'a}nchez, and J.~Piater, ``Continual learning from demonstration of robotics skills,'' \emph{Robotics and Autonomous Systems}, vol. 165, p. 104427, 2023.

\bibitem{EWC-1}
J.~Kirkpatrick, R.~Pascanu, N.~Rabinowitz, J.~Veness, G.~Desjardins, A.~A. Rusu, K.~Milan, J.~Quan, T.~Ramalho, A.~Grabska-Barwinska \emph{et~al.}, ``Overcoming catastrophic forgetting in neural networks,'' \emph{Proceedings of the national academy of sciences}, vol. 114, no.~13, pp. 3521--3526, 2017.

\bibitem{EWC-2}
J.~Schwarz, W.~Czarnecki, J.~Luketina, A.~Grabska-Barwinska, Y.~W. Teh, R.~Pascanu, and R.~Hadsell, ``Progress \& compress: A scalable framework for continual learning,'' in \emph{International conference on machine learning}.\hskip 1em plus 0.5em minus 0.4em\relax PMLR, 2018, pp. 4528--4537.

\bibitem{SI-1}
F.~Zenke, B.~Poole, and S.~Ganguli, ``Continual learning through synaptic intelligence,'' in \emph{International conference on machine learning}.\hskip 1em plus 0.5em minus 0.4em\relax PMLR, 2017, pp. 3987--3995.

\bibitem{RWALK}
A.~Chaudhry, P.~K. Dokania, T.~Ajanthan, and P.~H. Torr, ``Riemannian walk for incremental learning: Understanding forgetting and intransigence,'' in \emph{Proceedings of the European conference on computer vision (ECCV)}, 2018, pp. 532--547.

\bibitem{VCL-1}
\BIBentryALTinterwordspacing
C.~V. Nguyen, Y.~Li, T.~D. Bui, and R.~E. Turner, ``Variational continual learning,'' in \emph{International Conference on Learning Representations}, 2018. [Online]. Available: \url{https://openreview.net/forum?id=BkQqq0gRb}
\BIBentrySTDinterwordspacing

\bibitem{VI-1}
D.~M. Blei, A.~Kucukelbir, and J.~D. McAuliffe, ``Variational inference: A review for statisticians,'' \emph{Journal of the American statistical Association}, vol. 112, no. 518, pp. 859--877, 2017.

\bibitem{VI-2}
C.~Zhang, J.~B{\"u}tepage, H.~Kjellstr{\"o}m, and S.~Mandt, ``Advances in variational inference,'' \emph{IEEE transactions on pattern analysis and machine intelligence}, vol.~41, no.~8, pp. 2008--2026, 2018.

\bibitem{FRCL}
\BIBentryALTinterwordspacing
M.~K. Titsias, J.~Schwarz, A.~G. de~G.~Matthews, R.~Pascanu, and Y.~W. Teh, ``Functional regularisation for continual learning with gaussian processes,'' in \emph{International Conference on Learning Representations}, 2020. [Online]. Available: \url{https://openreview.net/forum?id=HkxCzeHFDB}
\BIBentrySTDinterwordspacing

\bibitem{GP}
E.~Schulz, M.~Speekenbrink, and A.~Krause, ``A tutorial on gaussian process regression: Modelling, exploring, and exploiting functions,'' \emph{Journal of mathematical psychology}, vol.~85, pp. 1--16, 2018.

\bibitem{FROMP}
P.~Pan, S.~Swaroop, A.~Immer, R.~Eschenhagen, R.~Turner, and M.~E.~E. Khan, ``Continual deep learning by functional regularisation of memorable past,'' \emph{Advances in neural information processing systems}, vol.~33, pp. 4453--4464, 2020.

\bibitem{Laplace-1}
A.~Kristiadi, M.~Hein, and P.~Hennig, ``Learnable uncertainty under laplace approximations,'' in \emph{Uncertainty in Artificial Intelligence}.\hskip 1em plus 0.5em minus 0.4em\relax PMLR, 2021, pp. 344--353.

\bibitem{S-FSVI-1}
T.~G. Rudner, F.~B. Smith, Q.~Feng, Y.~W. Teh, and Y.~Gal, ``Continual learning via sequential function-space variational inference,'' in \emph{International Conference on Machine Learning}.\hskip 1em plus 0.5em minus 0.4em\relax PMLR, 2022, pp. 18\,871--18\,887.

\bibitem{BNN-1}
L.~V. Jospin, H.~Laga, F.~Boussaid, W.~Buntine, and M.~Bennamoun, ``Hands-on bayesian neural networks—a tutorial for deep learning users,'' \emph{IEEE Computational Intelligence Magazine}, vol.~17, no.~2, pp. 29--48, 2022.

\bibitem{EKF}
S.~Julier and J.~K. Uhlmann, ``A general method for approximating nonlinear transformations of probability distributions,'' \emph{Citeseer}, 1996.

\bibitem{EnKF}
G.~Evensen, ``Sequential data assimilation with a nonlinear quasi-geostrophic model using monte carlo methods to forecast error statistic,'' \emph{JOURNAL OF GEOPHYSICAL RESEARCH}, vol.~99, no.~C5, pp. 10,143--10,162, 1994.

\bibitem{KL-Cauchy}
F.~Chyzak and F.~Nielsen, ``A closed-form formula for the kullback-leibler divergence between cauchy distributions,'' \emph{arXiv preprint arXiv:1905.10965}, 2019.

\bibitem{wasserstein-1}
V.~M. Panaretos and Y.~Zemel, ``Statistical aspects of wasserstein distances,'' \emph{Annual review of statistics and its application}, vol.~6, no.~1, pp. 405--431, 2019.

\bibitem{KL-1}
T.~Van~Erven and P.~Harremos, ``R{\'e}nyi divergence and kullback-leibler divergence,'' \emph{IEEE Transactions on Information Theory}, vol.~60, no.~7, pp. 3797--3820, 2014.

\bibitem{KL-Gaussian}
Y.~Zhang, J.~Pan, L.~K. Li, W.~Liu, Z.~Chen, X.~Liu, and J.~Wang, ``On the properties of kullback-leibler divergence between multivariate gaussian distributions,'' \emph{Advances in Neural Information Processing Systems}, vol.~36, 2024.

\bibitem{wasserstein-2}
C.~R. Givens and R.~M. Shortt, ``A class of wasserstein metrics for probability distributions.'' \emph{Michigan Mathematical Journal}, vol.~31, no.~2, pp. 231--240, 1984.

\bibitem{wassertein-3}
V.~D. Wild, R.~Hu, and D.~Sejdinovic, ``Generalized variational inference in function spaces: Gaussian measures meet bayesian deep learning,'' \emph{Advances in Neural Information Processing Systems}, vol.~35, pp. 3716--3730, 2022.

\bibitem{S-FSVI-2}
\BIBentryALTinterwordspacing
A.~Scannell, R.~Mereu, P.~E. Chang, E.~Tamir, J.~Pajarinen, and A.~Solin, ``Function-space parameterization of neural networks for sequential learning,'' in \emph{The Twelfth International Conference on Learning Representations}, 2024. [Online]. Available: \url{https://openreview.net/forum?id=2dhxxIKhqz}
\BIBentrySTDinterwordspacing

\bibitem{VAR-GP}
S.~Kapoor, T.~Karaletsos, and T.~D. Bui, ``Variational auto-regressive gaussian processes for continual learning,'' in \emph{International Conference on Machine Learning}.\hskip 1em plus 0.5em minus 0.4em\relax PMLR, 2021, pp. 5290--5300.

\bibitem{edge_device}
G.~Sivapalan, K.~K. Nundy, S.~Dev, B.~Cardiff, and D.~John, ``Annet: A lightweight neural network for ecg anomaly detection in iot edge sensors,'' \emph{IEEE Transactions on Biomedical Circuits and Systems}, vol.~16, no.~1, pp. 24--35, 2022.

\end{thebibliography}

\end{document}